\documentclass{acm-proc-article-sp}

\usepackage[margin=1in]{geometry}
\usepackage[font=footnotesize]{subfig}
\usepackage{graphicx}
\usepackage{verbatim}
\usepackage{hyperref}
\usepackage{url}
\usepackage{float}

\newcommand{\oled}{\textsf{\footnotesize OLED}}

\def\osla{$\mathtt{OSL}\alpha$}

\def\folholdsAt{\mathtt{HoldsAt}}
\def\folinitiatedAt{\mathtt{InitiatedAt}}
\def\folterminatedAt{\mathtt{TerminatedAt}}
\def\move{$\mathtt{move}$}
\def\meet{$\mathtt{meet}$}

\title{The Complex Event Recognition Group}
\author{Elias Alevizos \and Alexander Artikis \and Nikos Katzouris \and Evangelos Michelioudakis \and Georgios Paliouras\\ \affaddr{Institute of Informatics \& Telecommunications,} \\ \affaddr{National Centre for Scientific Research (NCSR) Demokritos, Athens, Greece}\\ \email{\{alevizos.elias, a.artikis, nkatz, vagmcs, paliourg\}@iit.demokritos.gr}}
\date{}

\begin{document}
\pagestyle{empty}

\maketitle

\begin{abstract}
The \emph{Complex Event Recognition} (CER) group is a research team,
affiliated with the National Centre of Scientific Research ``Demokritos'' in Greece.
The CER\linebreak  group works towards advanced and efficient methods for the recognition of complex events in a multitude of large, heterogeneous and interdependent data streams.
Its research covers multiple aspects of complex event recognition,
from efficient detection of patterns on event streams to
handling uncertainty and noise in streams, and machine learning techniques for inferring interesting patterns. Lately, it has expanded to methods for forecasting the occurrence of events.
It was founded in 2009 and currently hosts 3 senior researchers, 5 PhD students and works regularly with under-graduate students.
\end{abstract}

\section{Introduction}
\label{section:intro}

The proliferation of devices that work in real-time, constantly producing data streams, 
has led to a  paradigm shift with respect to what is expected from a system working with massive amounts of data. The dominant model for processing large-scale data was one that assumed a relatively fixed database/knowledge base, i.e., it assumed that the operations of updating existing records/facts and inserting new ones were infrequent. The user of such a system would then pose queries to the database, without very strict requirements in terms of latency.

While this model is far from being rendered obsolete (on the contrary), a system aiming to extract actionable knowledge from continuously evolving streams of data has to address a new set of challenges and satisfy a new set of requirements. The basic idea behind such a system is that it is not always possible, or even desirable, to store every bit of the incoming data, so that it can be later processed.
Rather, the goal is to make sense out of these streams of data, without having to store them. This is done by defining a set of queries/patterns, continuously applied to the data streams. Each such pattern includes a set of temporal constraints and, possibly, a set of spatial constraints, expressing a composite or complex event of special significance for a given application. The system must then be efficient enough so that instances of pattern satisfaction can be reported to a user with minimal latency. Such systems are called Complex Event Recognition (CER) systems \cite{DBLP:journals/csur/CugolaM12, DBLP:journals/pvldb/GiatrakosADG17,DBLP:journals/csur/AlevizosSAP17}.
  
CER systems are widely adopted in contemporary applications.  
Such applications are the recognition of attacks in computer network nodes, human activities on video content, emerging stories and trends on the Social Web, traffic and transport incidents in smart cities, fraud in electronic marketplaces, cardiac arrhythmias and epidemic spread. 
Moreover, Big Data frameworks, such as Apache Storm, Spark Streaming and Flink, have been extending their stream processing functionality by including implementations for  CER.

There are multiple issues that arise for a CER system. As already mentioned, one issue is the requirement for minimal latency. Therefore, a CER system has to employ highly efficient reasoning mechanisms, scalable to high-velocity streams. Moreover, pre-processing steps, like data cleaning, have to be equally efficient, otherwise they constitute a ``luxury'' that a CER system cannot afford. In this case, the system must be able to handle noise. This may be a requirement, even if perfectly clean input data is assumed, since domain knowledge is often insufficient or incomplete. Hence, the patterns defined by the users may themselves carry a certain degree of uncertainty. Moreover, it is quite often the case that such patterns cannot be provided at all, even by domain experts. This poses a further challenge of how to apply machine learning techniques in order to extract patterns from streams before a CER system can actually run with them. Standard machine learning techniques are not always directly applicable, due to the size and variability of the training set. As a result, machine learning techniques must work in an online fashion. Finally, one often needs to move beyond detecting instances of pattern satisfaction into forecasting when a pattern is likely to be satisfied in the future.

Our CER group\footnote{\url{http://cer.iit.demokritos.gr/}} at the National Centre for Scientific Research (NCSR) Demokritos, in Athens, Greece, has been conducting research on CER for the past decade, and has developed a number of novel algorithms and publicly available software tools. In what follows, we sketch the approaches that we have proposed and present some indicative  results.

\section{Complex Event Recognition}\label{sec:cer}

Numerous CER systems have been proposed in the literature \cite{DBLP:journals/csur/CugolaM12, DBLP:journals/pvldb/GiatrakosADG17}. 
Recognition systems with a logic-based representation of complex event (CE) patterns, in particular, have been attracting attention since they exhibit a formal, declarative semantics \cite{DBLP:journals/csur/AlevizosSAP17}. We have been developing an efficient dialect of the Event Calculus, called `Event Calculus for Run-Time reasoning' (RTEC) \cite{DBLP:journals/tkde/ArtikisSP15}. The Event Calculus is a logic programming formalism for representing and reasoning about events and their effects \cite{DBLP:journals/ngc/KowalskiS86}. CE patterns in RTEC identify the conditions in which a CE is initiated and terminated. Then, according to the law of inertia, a CE holds at a time-point $T$ if it has been initiated at some time-point earlier than $T$, and has not been terminated in the meantime. 

RTEC has been optimised for CER, in order to be scalable to high-velocity data streams. A form of caching stores the results of subcomputations in the computer memory to avoid unnecessary recomputations. A set of interval manipulation constructs simplify CE patterns and improve reasoning efficiency. A simple indexing mechanism makes RTEC robust to events that are irrelevant to the patterns we want to match and so RTEC can operate without data filtering modules. Finally, a `windowing' mechanism supports real-time CER. One main motivation for RTEC is that it should remain efficient and scalable in applications where events arrive with a (variable) delay from, or are revised by, the underlying sensors: RTEC can update the intervals of the already recognised CEs, and recognise new CEs, when data arrive with a delay or following revision. 

RTEC has been analysed theoretically, through a complexity analysis, and assessed experimentally in several application domains, including city transport and traffic management \cite{DBLP:conf/edbt/ArtikisWSBLPBMKMGMGK14}, activity recognition on video feeds \cite{DBLP:journals/tkde/ArtikisSP15}, and maritime monitoring \cite{DBLP:journals/geoinformatica/PatroumpasAAVPT17}.
In all of these applications, RTEC has proven capable of performing real-time CER, scaling to large data streams and highly complex event patterns.



\section{Uncertainty Handling}\label{section:cer_uncertainty}

CER applications exhibit various types of uncertainty, ranging from incomplete and erroneous data streams to imperfect CE patterns \cite{DBLP:journals/csur/AlevizosSAP17}. 
We have been developing techniques for handling uncertainty in CER by extending the Event Calculus with probabilistic reasoning.
Prob-EC \cite{SkarlatidisAFP15} is a logic programming implementation of the Event Calculus using the ProbLog engine \cite{DBLP:journals/tplp/KimmigDRCR11}, that incorporates probabilistic semantics into logic programming.  
Prob-EC is the first Event Calculus dialect able to deal with uncertainty in the input data streams.
For example, Prob-EC is more resilient to spurious data than the standard (crisp) Event Calculus.

\begin{figure}[t]
\centering
\includegraphics[width=0.48\textwidth]{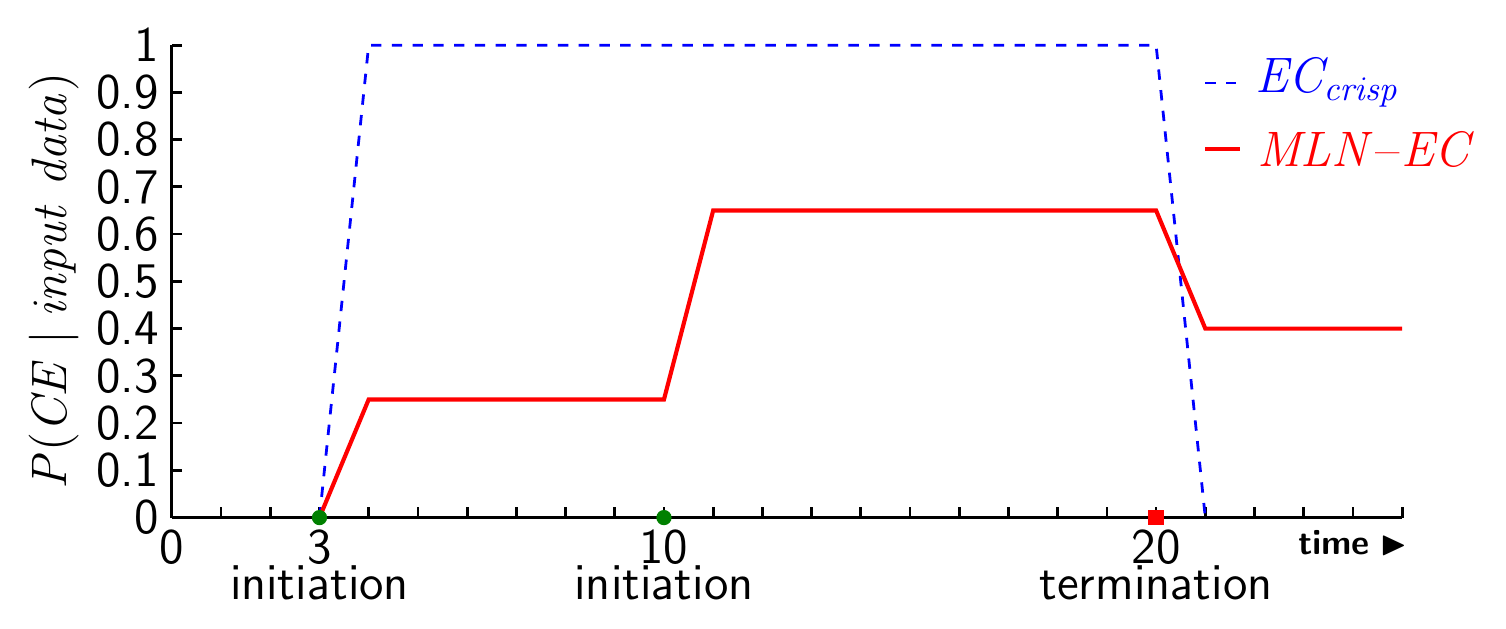}
\caption{CE probability estimation in the Event Calculus. The solid line concerns a probabilistic Event Calculus, such as MLN-EC, while the dashed line corresponds to a crisp (non-probabilistic) version of the Event Calculus. Due to the law of inertia, the CE probability remains constant in the absence of input data. Each time the initiation conditions are satisfied (e.g., in time-points 3 and 10), the CE probability increases. Conversely, when the termination conditions are satisfied (e.g., in time-point 20), the CE probability decreases.} \label{fig:probabilistic-EC}
\end{figure}

MLN-EC \cite{anskarlTOCL15} is an Event Calculus implementation based on Markov Logic Networks (MLN)s \cite{domingos2006markov}, a framework that combines first-order logic with graphical models, in order to enable probabilistic inference and learning. 
CE patterns may be associated with weight values, indicating our confidence in them. Inference can then be performed regarding the time intervals during which CEs of interest hold. 
Like Prob-EC, MLN-EC increases the probability of a CE every time its initiating conditions are satisfied, and decreases this probability whenever its terminating conditions are satisfied, as shown in Figure \ref{fig:probabilistic-EC}.
Moreover, in MLN-EC the domain-independent Event Calculus rules, expressing the law of inertia, may be associated with weight values, introducing probabilistic inertia. This way, the model is highly customisable, by tuning appropriately the weight values with the use of machine learning techniques, and thus achieves high predictive accuracy in a wide range applications. 


\begin{figure}[t]
\centering
\includegraphics[width=0.38\textwidth]{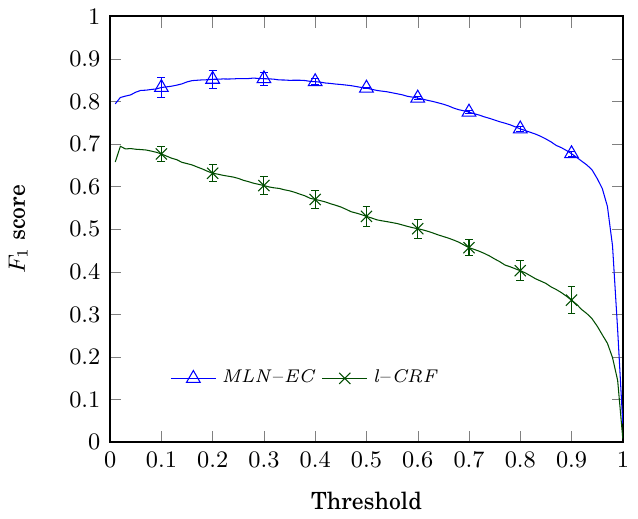}
\caption{CER under uncertainty. $F_1$-score of MLN-EC and linear-chain CRFs for different CE acceptance thresholds.} \label{fig:MLN-EC_vs_CRF}
\end{figure}


The use of background knowledge about the task and the domain, in terms of logic (the Event Calculus), can make MLN-EC more robust to variations in the data. Such variations are very common in practice, particularly in dynamic environments, such as the ones encountered in CER. The common assumption made in machine learning that the training and test data share the same statistical properties is often violated in these situations. 
Figure \ref{fig:MLN-EC_vs_CRF}, for example, compares the performance of MLN-EC against linear-chain Conditional Random Fields on a benchmark activity recognition dataset, where evidence is incomplete in the test set as compared to the training set.



\section{Event Pattern Learning}
\label{section:ml}

The manual authoring of CE patterns is a tedious and error-prone process. Consequently, the automated construction of such patterns from data is highly desirable. 
We have been developing supervised, online learning learning tools for constructing logical representations of CE patterns, from a single-pass over a relational data stream. 
%
\osla\ \cite{vagmcs2016osla} is such a learner for Markov Logic Networks (MLNs),  formulating CE patterns in the form of MLN-EC theories. \osla\ extends OSL \cite{huynh2011osl} by exploiting a background knowledge in order to significantly constrain the search for patterns.

In each step $t$ of the online procedure, a set of training examples $\mathcal{D}_t$ arrives containing input data along with CE annotation. $\mathcal{D}_t$ is used together with the already learnt hypothesis, if any, to predict the truth values of the CEs of interest. This is achieved by MAP (maximum a posteriori) inference. Given $\mathcal{D}_t$, \osla\ constructs a hypergraph that represents the space of possible structures as graph paths. Then, for all incorrectly predicted CEs, the hypergraph is searched using relational pathfinding, for clauses supporting the recognition of these CEs. The paths discovered during the search are generalised into first-order clauses. Subsequently, 
the weights of the clauses that pass the evaluation stage are optimised using off-the-shelf online weight learners. 
Then, the weighted clauses are appended to the hypothesis and the procedure is repeated for the next set of training examples $\mathcal{D}_{t+1}$.

\oled \ \cite{DBLP:journals/tplp/KatzourisAP16} is an Inductive Logic Programming (ILP) system that learns CE patterns, in the form of Event Calculus theories, in a supervised fashion and in a single pass over a data stream. 
\oled \ constructs patterns by first encoding a positive example from the input stream into a so-called \emph{bottom rule}, i.e., a most-specific rule of the form $\alpha \leftarrow \delta_1 \wedge \ldots \wedge \delta_n$, where $\alpha$ is an initiation or termination atom and $\delta_1, \ldots, \delta_n$ are relational features, constructed on-the-fly, using prescriptions from a predefined \emph{language bias}. These features express anything ``interesting'',  as defined by the language bias, that is true within the positive example at hand. A bottom rule is typically too restrictive to be useful. To learn a useful rule, \oled \ searches within the space of rules that $\theta$-\emph{subsume} the bottom rule, i.e., rules that involve some of the $\delta_i$'s only. To that end, \oled \ starts from the most-general rule---a rule with an empty body---and gradually \emph{specialises} that rule by adding $\delta_i$'s to its body, using a rule evaluation function to assess the quality of each generated specialisation.

\oled's single-pass strategy is based on the \emph{Hoeffding bound} \cite{hoeffding1963probability}, a statistical tool that allows to approximate the quality of a rule on the entire input using only a subset of the data. In particular, given a rule $r$ and some of its specialisations $r_1, \ldots, r_k$, the Hoeffding bound allows to identify the best among them, with probability $1{-}\delta$ and within an error margin $\epsilon$, using only $N = \mathcal{O}(\frac{1}{\epsilon^2}ln\frac{1}{\delta})$ training examples from the input stream. 

We have evaluated \oled\ and \osla\ on real \linebreak datasets concerning activity recognition, maritime monitoring, credit card fraud detection, and traffic management in smart cities \cite{DBLP:journals/tplp/KatzourisAP16, vagmcs2016osla, DBLP:conf/debs/ArtikisKCBMSFP17, michelioudakis2016online, nkatzILP2017}. 
We have also compared \oled\ and \osla\ to OSL \cite{huynh2011osl}, XHAIL, a `batch' structure learner requiring many passes over the data \cite{Ray09}, and to hand-curated Event Calculus patterns (with optimised weight values).
The results suggest that both \oled\ and \osla\ can match the predictive accuracy of batch learners as well as that of hand-crafted patterns. Moreover, \oled\ and \osla\ have proven significantly faster than both batch and online learners, making them more appropriate for large data streams.

\section{Event forecasting}
\label{section:forecasting}

Forecasting over time-evolving data streams is a task that can be defined in multiple ways. 
There is a  conceptual difference between forecasting and prediction,
as the latter term is understood in machine learning,
where the main goal is to ``predict'' the output of a function on previously unseen input data,
even if there is no temporal dimension.
In forecasting, \emph{time} is a crucial component and the goal is to predict the temporally future output of some function or the occurrence of an event.
Time-series forecasting is an example of the former case and is a field with a significant history of contributions. However, its methods cannot be directly transferred to CER, since it handles streams of \linebreak (mostly) real-valued variables and focuses on forecasting relatively simple patterns. On the contrary, in CER we are also interested in categorical values, related through complex patterns and involving multiple variables. 
%
Our group has developed a method, where automata and Markov chains are employed in order to provide (future) time intervals during which a match is expected with a probability above a confidence threshold \cite{alevizos2017event}.

We start with a given pattern of interest, defining relations between events, in the form of a regular expression---i.e., using operators for \textit{sequence}, \textit{disjunction} and \textit{iteration}. Our goal, besides detecting occurrences of this pattern, 
is also to estimate, 
at each new event arrival, 
the number of future events that we will need to wait for until the expression is satisfied, and thus a match be detected.
A pattern in the form of a regular expression is first converted to a deterministic finite automaton (DFA) through standard conversion algorithms.
We then construct a Markov chain that will be able to provide a probabilistic description of the DFA's run-time behavior,
by employing Pattern Markov Chains (PMC) \cite{nuel_pattern_2008}. 
The resulting PMC depends both on the initial pattern and on the assumptions made about the statistical properties of the input stream---the order $m$ of the assumed Markov process. 

\begin{figure}[t]
\begin{centering}
\subfloat[Deterministic Finite Automaton, state 1.]{ 
\includegraphics[width=0.4\textwidth]{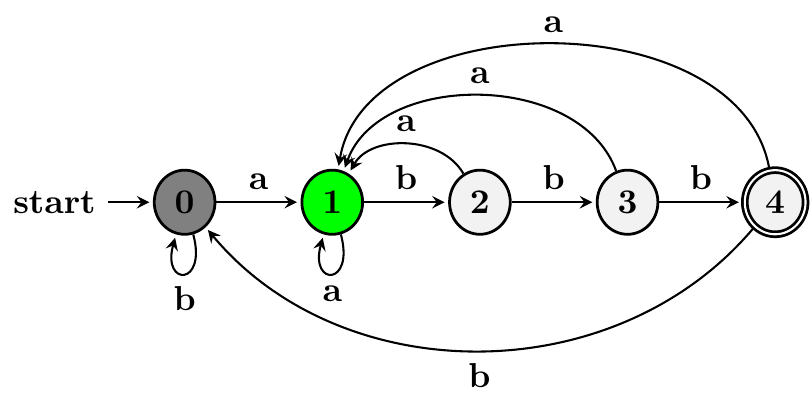}
\label{fig:dfa1}
}
\hfill
\subfloat[Waiting-time distribution, state 1.]{ 
\includegraphics[width=0.4\textwidth]{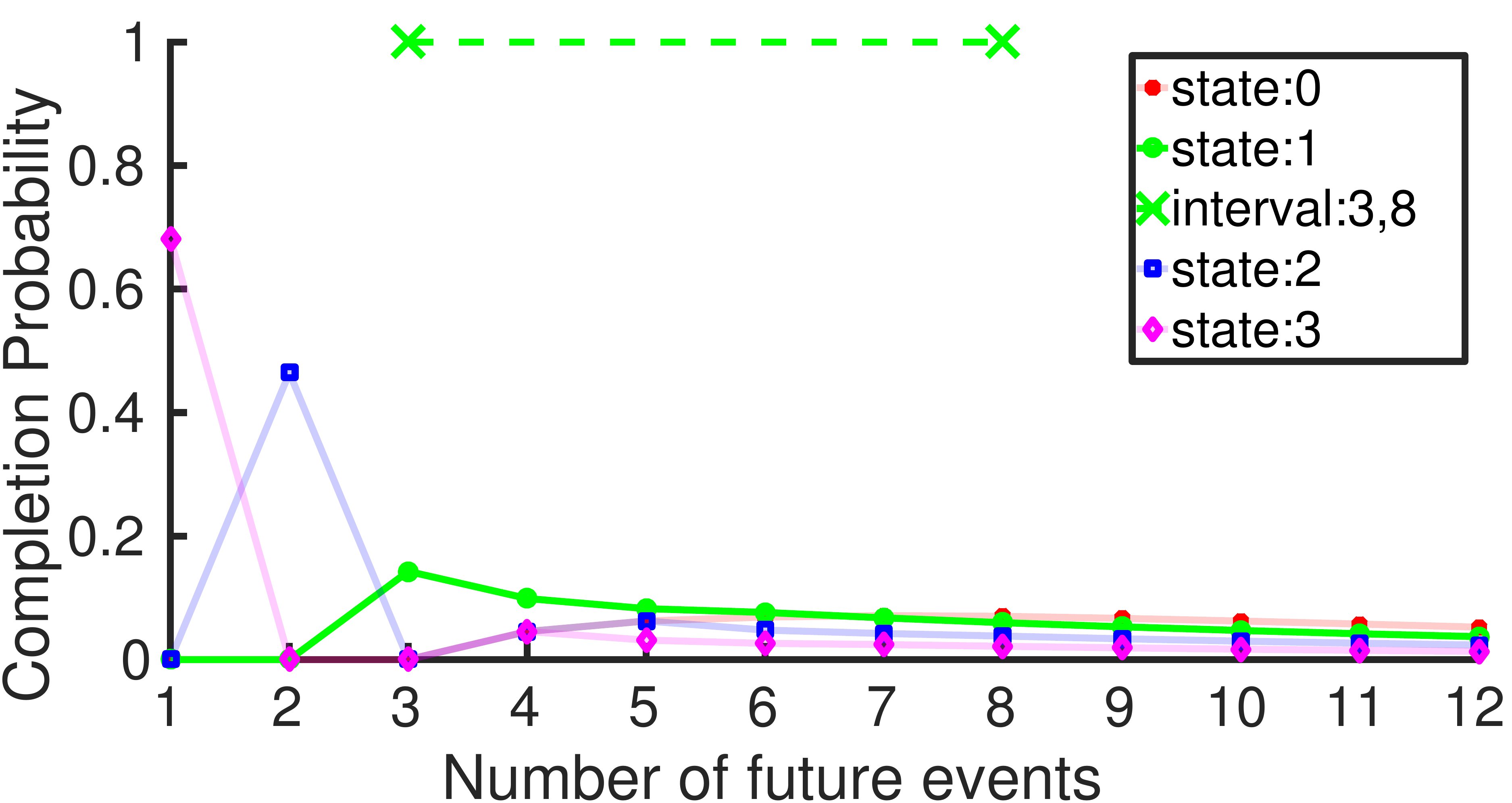}
\label{fig:wt1}
}
\caption{Event Forecasting. The event pattern requires that one event of type $a$ is followed by three events of type $b$. 
$\theta_{fc}=0.5$. For illustration, the $x$ axis stops at 12 future events.}
\label{fig:wtdfas}
\end{centering}
\end{figure}

After constructing a PMC, we can use it to calculate the so-called \textit{waiting-time} distributions, which can give us the probability of reaching a final state of the DFA in $k$ transitions from now. To estimate the final forecasts, another step is required, since our aim is not to provide a single future point with the highest probability, but an interval in the form of $I{=}(\mathit{start},\mathit{end})$. The meaning of such an interval is that the DFA is expected to reach a final state sometime in the future between $\mathit{start}$ and $\mathit{end}$ with probability at least some constant threshold $\mathit{\theta_{fc}}$ (provided by the user). 
An example is shown in Figure \ref{fig:wtdfas}, where the DFA in Figure \ref{fig:dfa1} is in state $1$, the \textit{waiting-time} distributions for all of its non-final states are shown in Figure \ref{fig:wt1}, and the distribution, along with the forecast interval, for state $1$ are shown in green.

\begin{figure*}[t]
\begin{centering}
\subfloat[Precision.]{ 
\includegraphics[width=0.33\textwidth,height=0.15\textheight]{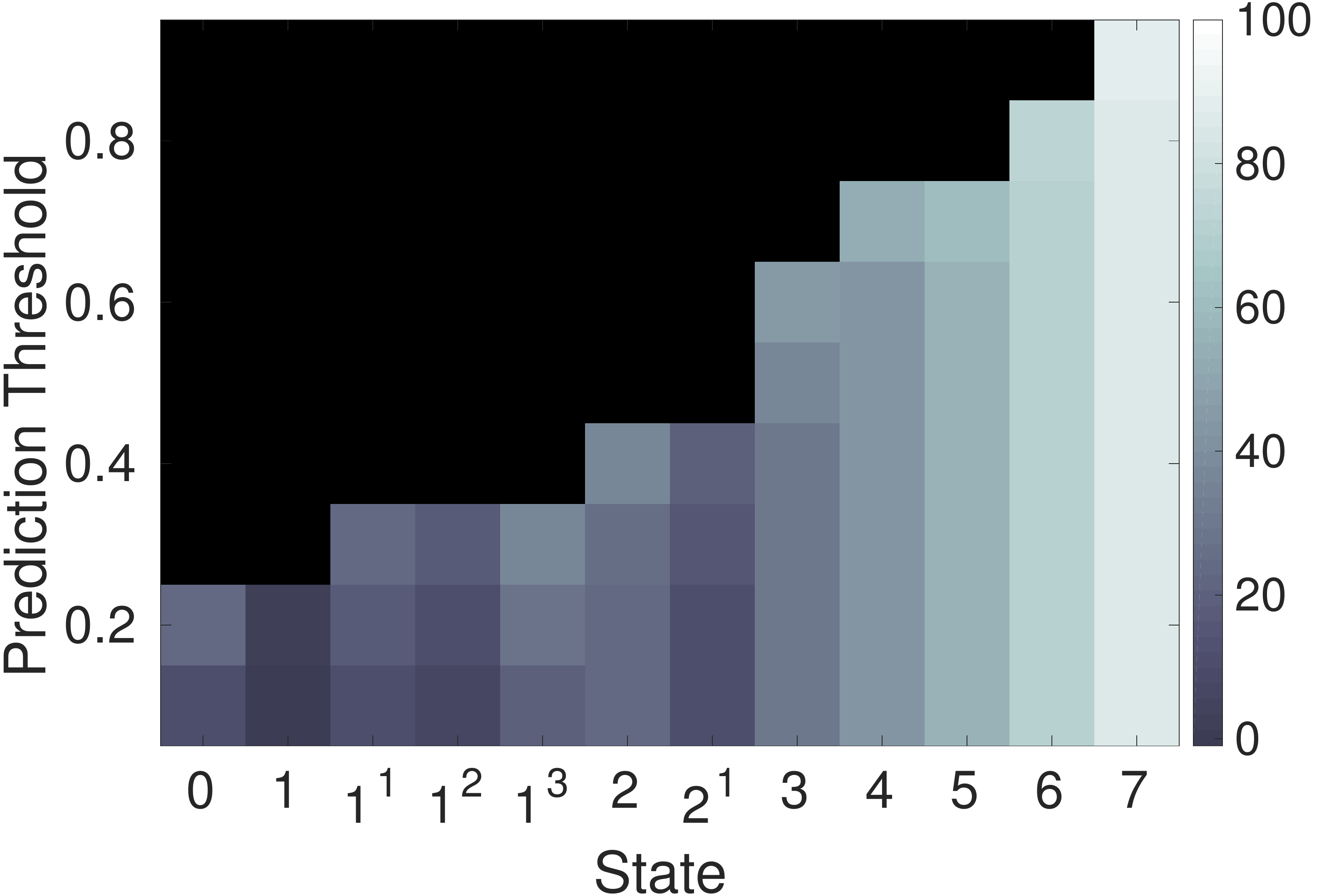}
\label{fig:fraud_3_precision}
}
\subfloat[Spread.]{ 
\includegraphics[width=0.33\textwidth,height=0.15\textheight]{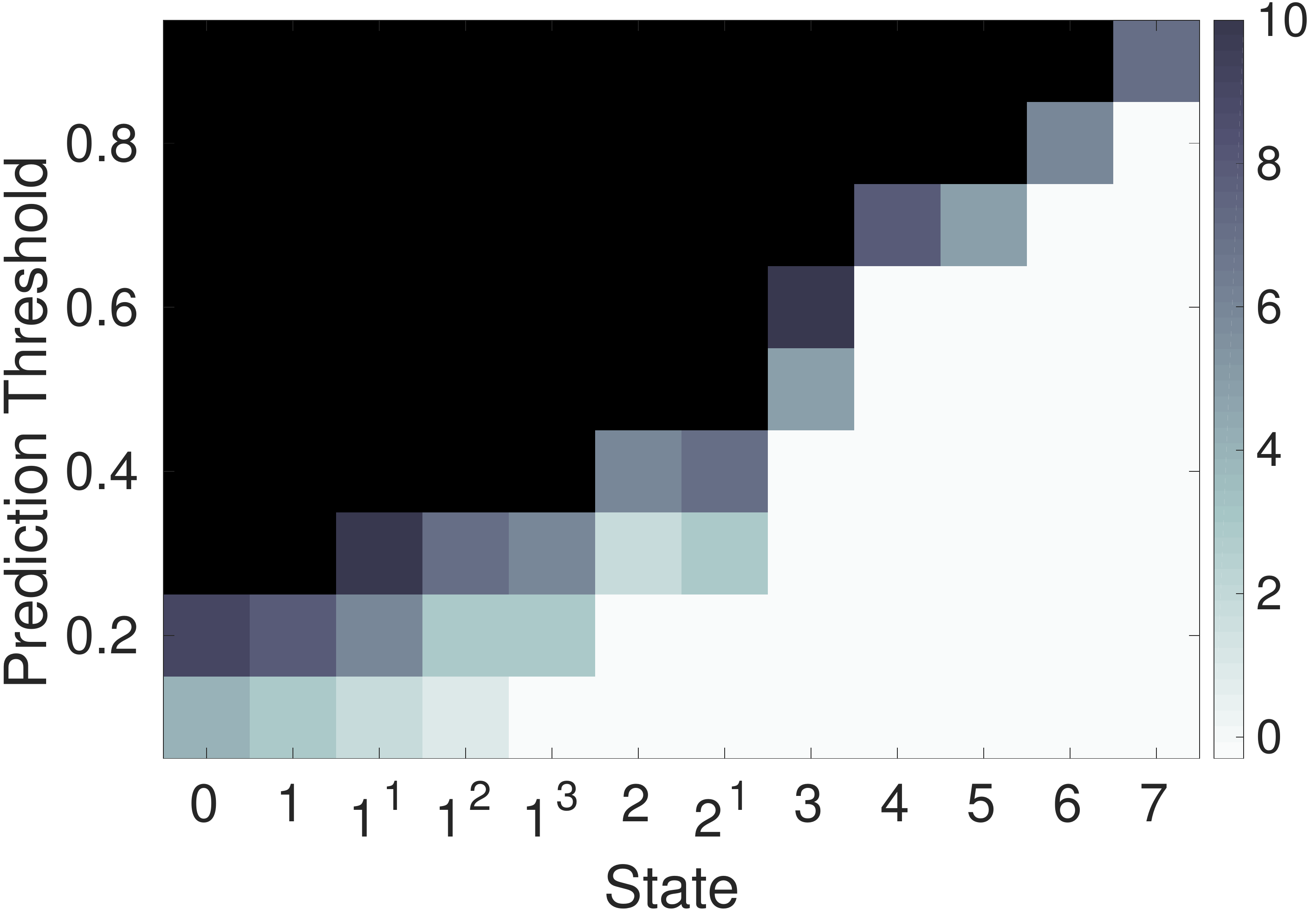}
\label{fig:fraud_3_spread}
}
\subfloat[Distance.]{ 
\includegraphics[width=0.33\textwidth,height=0.15\textheight]{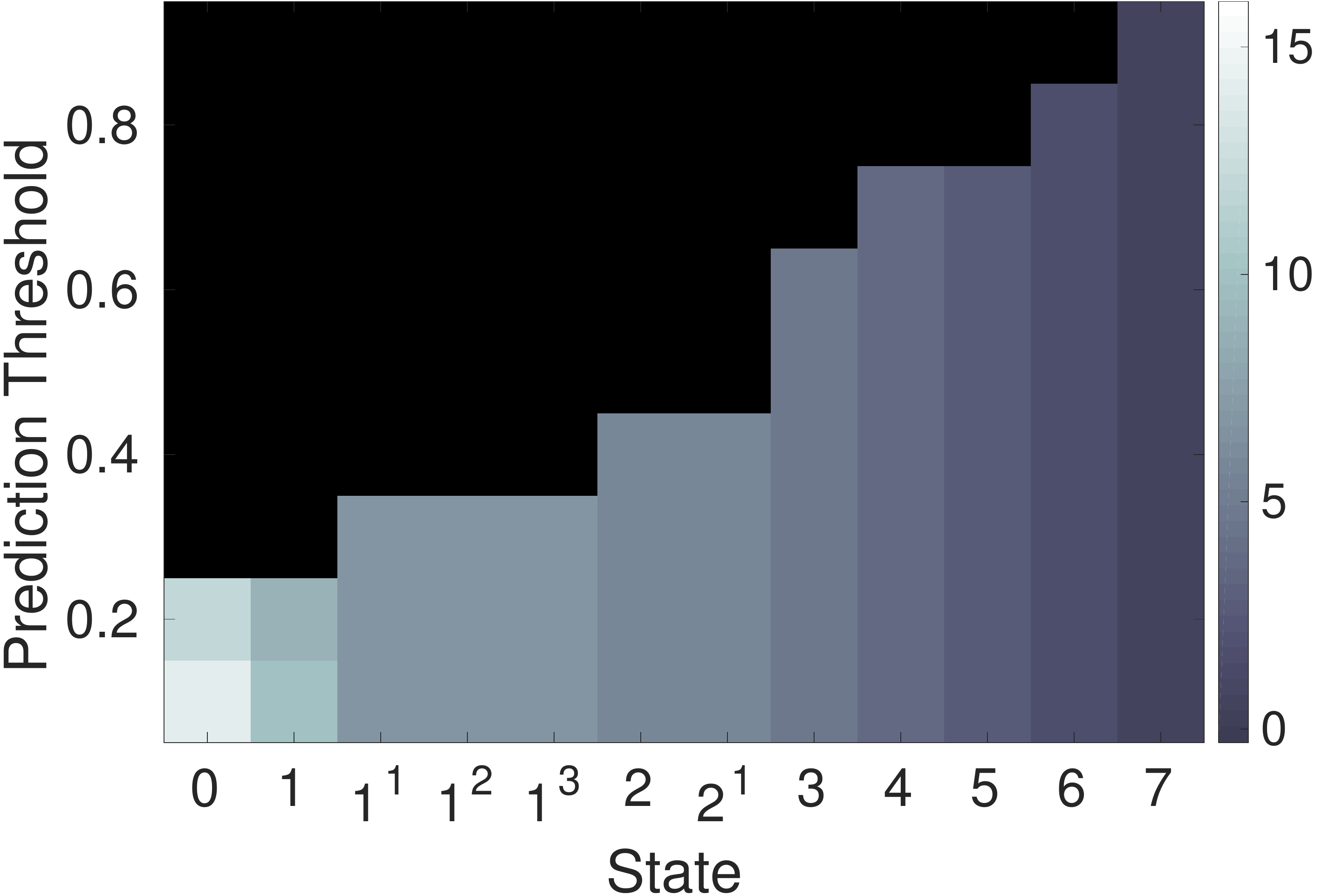}
\label{fig:fraud_3_distance}
}
\hfill
\subfloat[Precision.]{ 
\includegraphics[width=0.33\textwidth,height=0.15\textheight]{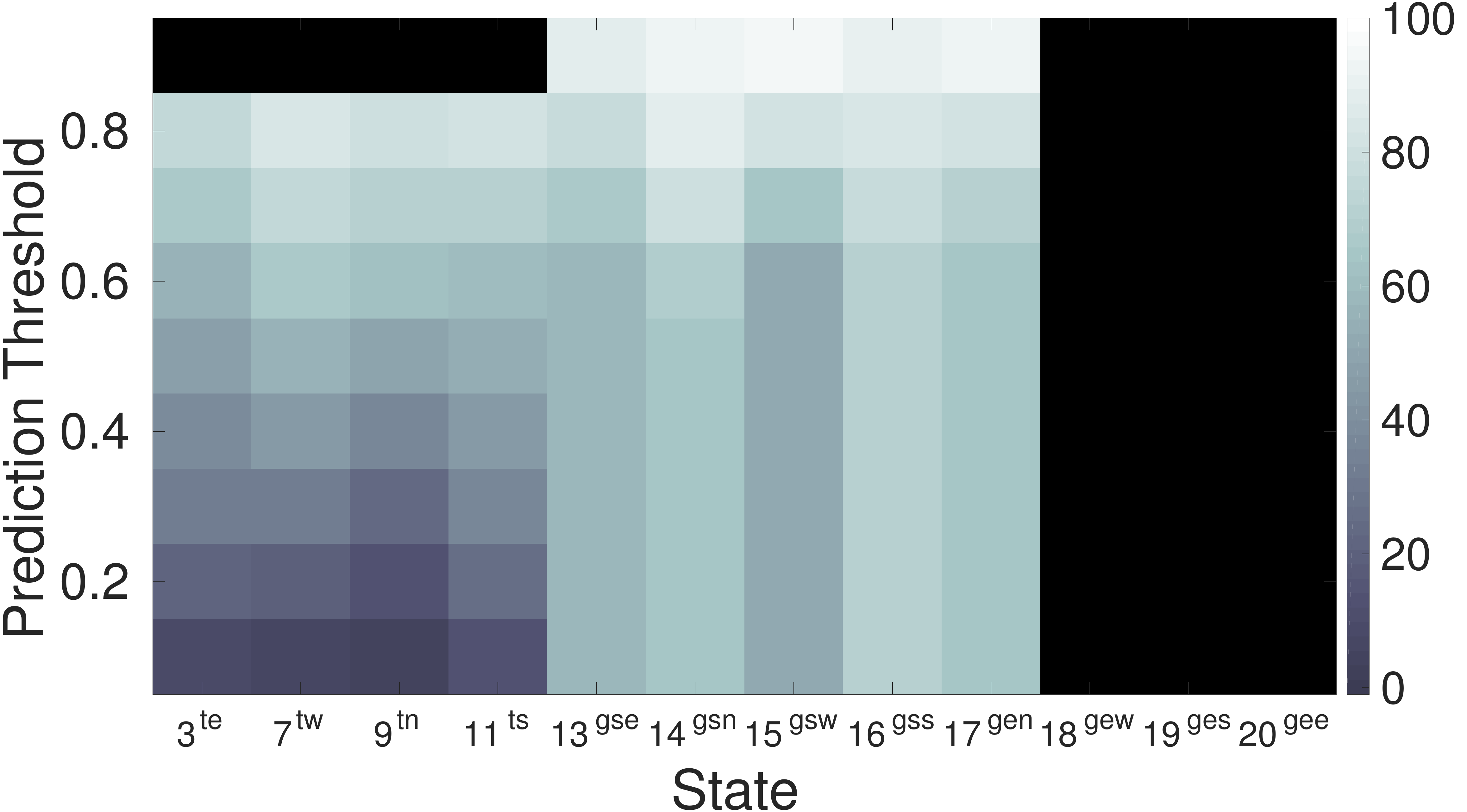}
\label{fig:maritime_1_precision}
}
\subfloat[Spread.]{ 
\includegraphics[width=0.33\textwidth,height=0.15\textheight]{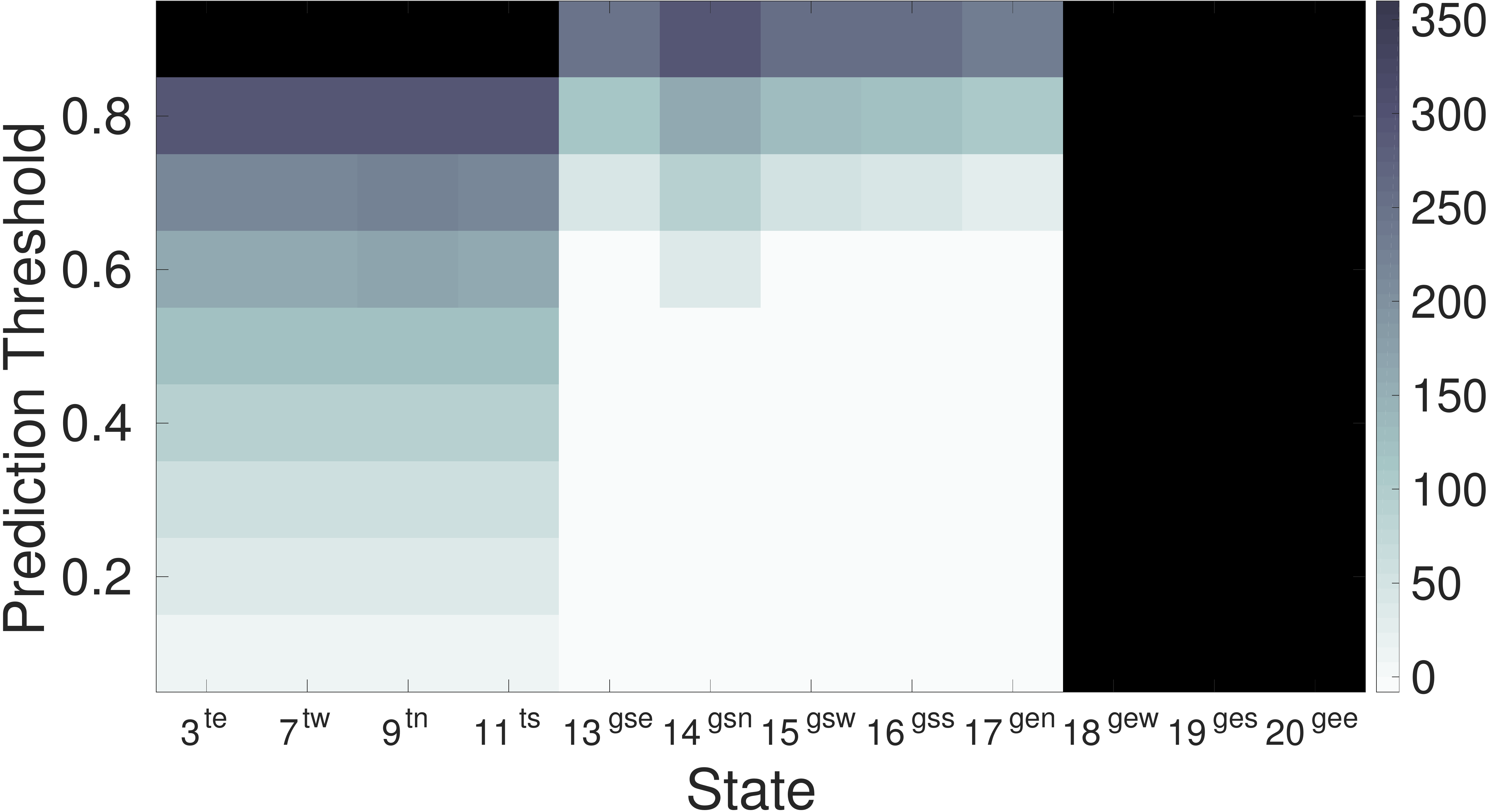}
\label{fig:maritime_1_spread}
}
\subfloat[Distance.]{ 
\includegraphics[width=0.33\textwidth,height=0.15\textheight]{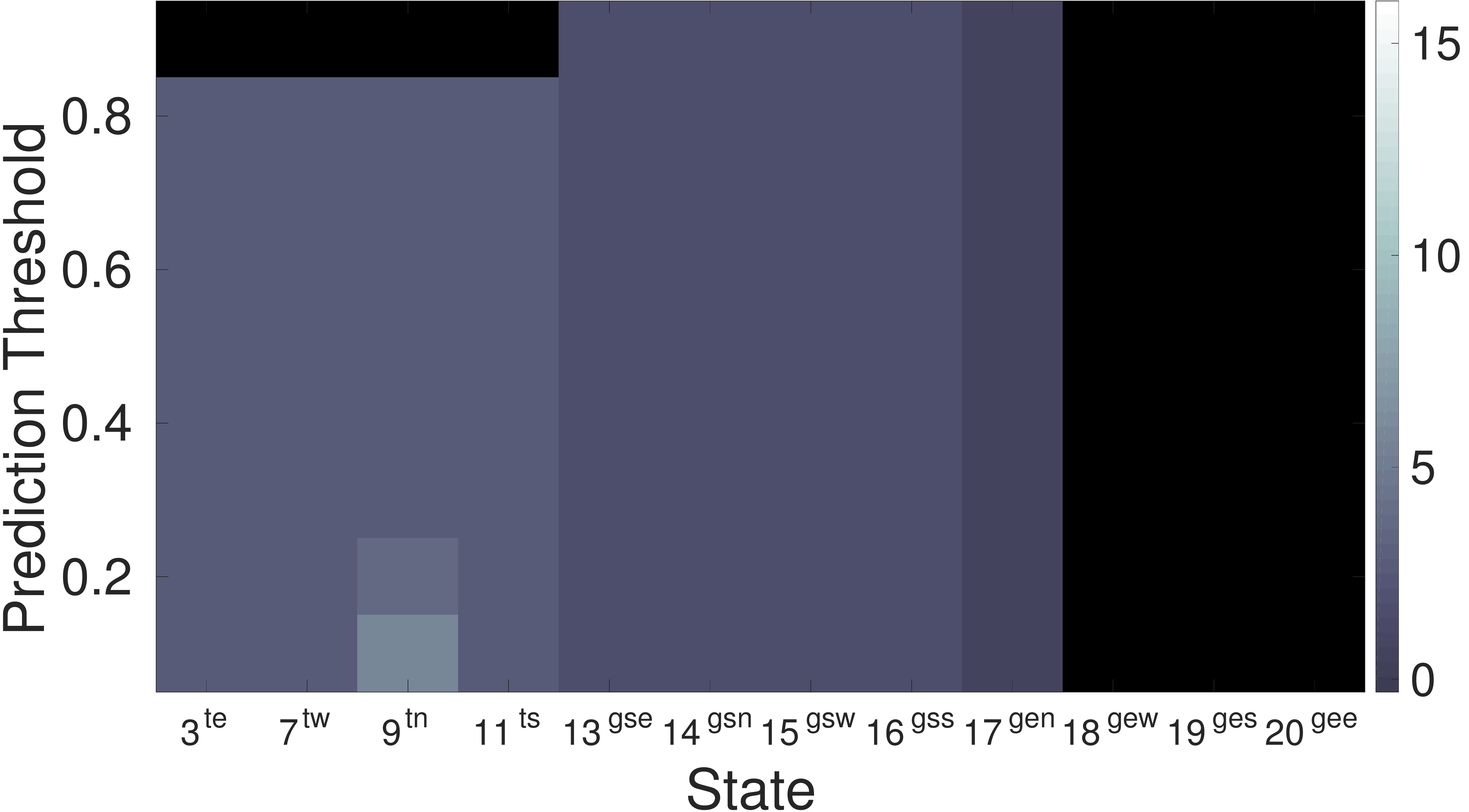}
\label{fig:maritime_1_distance}
}
\caption{Event forecasting for credit card fraud management (top) and maritime monitoring (bottom). 
The $y$ axes correspond to different values of the threshold $\mathit{\theta_{fc}}$. 
The $x$ axes correspond to states of the PMC.}
\label{fig:forecasting_results}
\end{centering}
\end{figure*}

Figure \ref{fig:forecasting_results} shows results of our implementation on two real-world datasets from the financial and the maritime domains.
In the former case, the goal was to forecast a specific case of credit card fraud, whereas in the latter it was to forecast a specific vessel manoeuver.
Figures \ref{fig:fraud_3_precision} and \ref{fig:maritime_1_precision} show \emph{precision} results (the percentage of forecasts that were accurate), where the $y$ axes correspond to different values of the threshold $\mathit{\theta_{fc}}$, and the $x$ axes correspond to states of the PMC 
(more ``advanced'' states are to the right of the axis), i.e., we measure precision for the forecasts produced by each individual state.
Similarly, Figures \ref{fig:fraud_3_spread} and \ref{fig:maritime_1_spread} are per-state plots for \emph{spread} (the length of the forecast interval), and Figures \ref{fig:fraud_3_distance} and \ref{fig:maritime_1_distance} are per-state plots for \emph{distance} (the temporal distance between the time a forecast is produced and the start of the forecast interval).

As expected, more ``advanced'' states produce forecasts with higher precision, smaller spread and distance.
However, there are cases where we can get earlier both high precision and low spread scores 
(see Figures \ref{fig:maritime_1_precision} and \ref{fig:maritime_1_spread}). 
This may happen when there exist strong probabilistic dependencies in the stream, e.g., when one event type is very likely (or very unlikely) to appear, given that the last event(s) is of a different event type. Our system can take advantage of such cases in order to produce high-quality forecasts early.

\section{Participation in Research \& Innovation Projects}
\label{sec:RIA}

The CER group has been participating in several research and innovation projects, contributing to the development of intelligent systems in challenging domains. 
\textbf{SPEEDD}\footnote{\url{http://speedd-project.eu/}} (Scalable Proactive Event-Driven Decision Making) was an FP7 EU-funded project, coordinated by the CER group, that developed tools for proactive analytics in Big Data applications. In SPEEDD, the CER group worked on credit card fraud detection and traffic management \cite{DBLP:conf/debs/ArtikisKCBMSFP17, michelioudakis2016online}, developing formal tools for highly scalable CER \cite{DBLP:journals/tkde/ArtikisSP15}, and pattern learning \cite{DBLP:journals/ml/KatzourisAP15, vagmcs2016osla}. 

\textbf{REVEAL}\footnote{\url{http://revealproject.eu/}} (REVEALing hidden concepts in social media) was an FP7 EU project that developed techniques for real-time extraction of knowledge from social media, including influence and reputation assessment. In REVEAL, the CER group developed a technique for \emph{online} (single-pass) learning of event patterns under uncertainty \cite{DBLP:journals/tplp/KatzourisAP16}.  

\textbf{datACRON}\footnote{\url{http://www.datacron-project.eu/}} (Big Data Analytics for Time Critical Mobility Forecasting) is an H2020 EU project that introduces novel methods for detecting threats and abnormal activity in very large fleets of moving entities, such as vessels and aircrafts. 
Similarly, \textbf{AMINESS}\footnote{\url{http://aminess.eu/}} (Analysis of Marine Information for Environmentally Safe Shipping) was a national project that developed a computational framework for environmental safety and cost reduction in the maritime domain. 
The CER group has been working on maritime and aviation surveillance, developing algorithms for, among others, highly efficient spatio-temporal pattern matching \cite{DBLP:journals/geoinformatica/PatroumpasAAVPT17}, complex\linebreak event forecasting \cite{alevizos2017event}, and parallel online learning of complex event patterns \cite{nkatzILP2017}.

\textbf{Track \& Know} (Big Data for Mobility \& Tracking Knowledge Extraction in Urban Areas) is an  H2020 EU-funded project that will research, develop and exploit a new software framework increasing the efficiency of Big Data applications in the transport, mobility, motor insurance and health sectors. The CER team is responsible for the complex event recognition and forecasting technology that will be developed in Track \& Know.


\section{Contributions to the community}
\label{section:community_my_ass}

The CER group supports the research community at different levels; notably, by making available the proposed research methods as open-source solutions. 
The RTEC CER engine (see Section \ref{sec:cer}) is available as a monolithic Prolog implementation\footnote{\url{https://github.com/aartikis/RTEC}} and as a parallel Scala implementation\footnote{\url{https://github.com/kontopoulos/ScaRTEC}}. 
The \oled\ system for online learning of event patterns (see Section \ref{section:ml}) is also available as an open-source solution\footnote{\url{https://github.com/nkatzz/OLED}}, both for single-core and parallel learning. \oled\ is implemented in Scala; both \oled\ and RTEC use the \textsf{Akka} actors library for parallel processing.

The \osla\ online learner (see Section \ref{section:ml}), along with MAP inference based on integer linear programming, and various weight optimisation algorithms (Max-Margin, CDA and AdaGrad), are contributed to LoMRF\footnote{\url{https://github.com/anskarl/LoMRF}}, an open-source implementation of Markov Logic Networks. LoMRF provides predicate completion, clausal form transformation, and function elimination. Moreover, it provides a parallel grounding algorithm which efficiently constructs the minimal Markov Random Field.  


\bibliographystyle{abbrv}
\bibliography{cer,forecasting,ml}

\end{document}